\documentclass[twoside,11pt]{article}

\usepackage{jmlr2e}
\usepackage{microtype}
\usepackage{graphicx}
\usepackage{subfigure}
\usepackage{booktabs} 
\usepackage{amssymb}
\usepackage{mathrsfs}
\usepackage{amsmath}
\usepackage{enumitem}
\usepackage{color}
\usepackage{hyperref}
\usepackage{natbib}
\usepackage{dsfont}
\usepackage{enumitem}
\usepackage{bbm}
\usepackage{algorithm}
\usepackage{algorithmic}

\usepackage{pgfplots}
\pgfplotsset{compat=newest}
\usetikzlibrary{plotmarks}
\usepackage{grffile}
\usetikzlibrary{plotmarks}
\usepackage{grffile}

\newcommand\argmax{\mathop{\rm arg\,max}}

\newcommand {\defn} {\triangleq}

\newcommand \E {\mathop{\mbox{\ensuremath{\mathbb{E}}}}\nolimits}

\newcommand \bel {\xi}

\newcommand{\cset}[2]{\left\{\, #1 ~\middle|~ #2 \,\right\} }

\newcommand{\Normal} {\mathcal{N}}

\newcommand {\pdf} {p}

\newcommand {\act} {x}
\newcommand {\obs} {y}
\newcommand {\Act} {X}
\newcommand {\Obs} {Y}
\newcommand {\rew} {r}
\newcommand {\rewf} {\rho}
\newcommand {\param} {\theta}
\newcommand {\Param} {\Theta}

\newcommand {\xt} {\act_t}
\newcommand {\bt} {\bel_t}
\newcommand {\cb}[2] {\xi_{#1 \mid #2}}
\newcommand {\cbt}[1] {\xi_{t \mid #1}}
\newcommand {\yt} {y_t}
\newcommand {\rt} {r_t}

\newcommand {\val} {V}
\newcommand {\qval} {Q}

\newcommand {\iregret} {R_T}
\newcommand {\aregret} {\bar{R}_T}
\newcommand {\sregret} {R^*_T}

\newcommand {\hV} {\hat{\val}}
\newcommand {\hVt} {\hV_t}

\newcommand {\Vt} {\val_t}
\newcommand {\VT} {\val_T}
\newcommand {\Qt} {\qval_t}

\newcommand {\hQt} {\hat{\qval}_t}

\newcommand {\depth} {d} 

\newcommand {\nThompson} {N}
\newcommand {\nMC} {M}

\newcommand \dd{\, \mathrm{d}}

\ShortHeadings{Drug discovery via Bayesian sparse sampling}{ERIKSSON, DIMITRAKAKIS AND CARLSSON}
\firstpageno{1}

\begin{document}

\title{High-dimensional near-optimal experiment design for drug discovery via Bayesian sparse sampling}

\author{\name Hannes Eriksson \email hannese@chalmers.se \\
       \addr Chalmers University of Technology, Gothenburg, Sweden\\
       \AND
		\name Christos Dimitrakakis\\
		\addr Chalmers University of Technology, Gothenburg, Sweden\\
		\AND
		\name Lars Carlsson\\
		\addr AstraZeneca AB, Gothenburg, Sweden
	}


\maketitle

\begin{abstract}
 We study the problem of performing automated experiment design for
drug screening through Bayesian inference and optimisation. In
particular, we compare and contrast the behaviour of linear-Gaussian
models and Gaussian processes, when used in conjunction with upper
confidence bound algorithms, Thompson sampling, or bounded horizon
tree search. We show that non-myopic sophisticated exploration
techniques using sparse tree search have a distinct advantage over
methods such as Thompson sampling or upper confidence bounds in this
setting. We demonstrate the significant superiority of the approach
over existing and synthetic datasets of drug toxicity. 
\end{abstract}

\section{Introduction}\label{sec:introduction}
We consider the problem of optimal adaptive experiment design in high-dimensional spaces. 
Informally, this is the problem of designing an
adaptive policy for performing a sequence of experiments, so as to
validate one or more hypothesis. For example, what sequence of observations should an astronomer take to best detect inhabitable planets? How should experimental treatments be adaptively allocated to patients so as to minimise adverse side effects and maximise the chances of discovering the best one?

In this paper, we are primarily interested in finding one molecule
that has optimal characteristics as defined by the read out from
experiments conducted with this molecule. The idea is to mimic the
drug-discovery process until the point where a candidate drug is
selected. A candidate drug is a molecule that meets certain criteria;
it needs to be transported to the therapeutic target, it needs to
affect the target in a way that is good for patients and it can not
cause the patients any other harm.

We adopt the setting of sequential experimentation, whereby we test
one or more drugs, observe their effects and update our beliefs, and
then perform another test.  This can be formalised as a type of bandit
problem~\cite{Degroot:OptimalStatisticalDecisions,chernoff1966smc,Chernoff:SequentialDesignExperiments,Lai+Robbins},
where at time $t$ the decision maker takes an action $\act_t \in
\Act$, which corresponds to performing a specific experiment, observes
a result $\obs \in \Obs$, which corresponds to obtaining a
measurement, and obtains a reward $\rt$. The decsion maker is
interested in maximising the utility $U = \sum_{t=1}^T \rt$, defined
as the total reward until the end of the game $T$.  We adapt the same
general framework, but typically the rewards for $t < T$ are small
negative values that reflect the cost of experimentation. However, the
reward at the last stage, $r_T$, is chosen to reflect the usefulness
of the information collected so far.We define the reward more
precisely in Section~\ref{sec:setting}. 

\subsection{Setting}
\label{sec:setting}
We adopt the standard Bayesian setting where we have a candidate
family of (conditional) densities $\{\pdf_\param \mid \theta \in
\Param\}$, parameterised by $\param$ . Each $\pdf_\param(\obs \mid
\act)$ defines a density over $\Obs$ for every $\act \in \Act$. If
$\param$ is known, then $\pdf_\param$ describes everything known about
the problem. However, since we do not know $\param$, we first select
some distribution $\bel_0$ on $\Param$, representing our prior
\emph{belief} about the unknown parameter. Through Bayesian updating,
we condition on the evidence, so that after $t$ experiments and
observations we obtain a new belief $\bel_t$ on the family indexed by
$\Param$.

In particular, in our setting we take a sequence of actions
$\act^t = \act_1, \ldots, \act_t$ with $\xt \in \Act$ and obtain a
corresponding sequence of outcomes $\obs^t = \obs_1, \ldots, \obs_t$
with $\yt \in \Obs$. As is standard, we define the posterior
belief at time $t$ to be the probability measure
\begin{align}
\label{eq:posterior belief}
\bel_t(B) &\defn
\cb{0}{\act^t, \obs^t}(B) \defn
\bel_0( B \mid \act^t, \obs^t)
\\
&=
\frac{\int_B \pdf_\theta(\obs^t \mid \act^t) \dd \bel_0(\theta)}{\pdf_{\bel_0}(\obs^t \mid \act^t)},
& B \subset \Theta,
\end{align}
where we use $\pdf_{\bel_0}(\obs^t \mid \act^t) = \int_\Theta \pdf_\theta(\obs^t \mid \act^t) \dd \bel_0(\theta)$ to denote the marginal density under $\bel_0$ and also introduce the convenient notation $\cb{0}{\act^t, \obs^t}$ to denote a particular belief conditioned on specific observations.

In our case in particular, we are performing Gaussian Process (GP)
inference. Thus, $\Param$ indexes a function space $F =
\cset{f_\param}{\param \in \Param}$ and we assume that our
observations follow a Gaussian distribution with variance $\sigma^2$
around the mean function $f$, i.e. that
\begin{equation}
\label{eq:gaussian-process}
\obs = f_\param(\act) + \Normal(0, \sigma),
\end{equation}
with expected value $\E_\param(\obs \mid \act) = f_\param(\act)$. 
In addition, any belief $\bel$ defines the corresponding expectation
$\E_\bel(\obs \mid \act) = \int_\Param f_\param(\act) \dd \bel(\param)$. 

We are interested in maximising the utility in expectation, under our
belief at each step. Here, we assume that the utility is an additive
function, is defined as
\[
U_t = \sum_{k=t}^T \rt,
\]
with rewards $\rt = \rewf_t(\xt, \yt, \bt)$ that can depend on the
current time, action, outcome and belief. The choice of reward
function is problem-dependent, but in the experiment design setting
there are thee standard choices: (a) When there is an inherent reward
for each observed outcome $\rewf(\yt)$, in which case we can simply set
$\rewf_t(\xt, \yt, \bt) = \rewf(\yt)$. This is the usual bandit setting.
(b) When we wish to find the best arm in the set as efficiently as possible.
This can be modelled
as $\rt =  - c(\xt)$, where $c$ describes the cost of taking a particular action.
for $t < T$, while the final reward is the expected value of the best arm
\[
r_T = \max_x \E_{\bt}(\rewf(\yt) \mid \xt = x).
\]
(c) The final choice is to try to learn as much as possible about the correct parameter. 
This can be modelled through maximising the KL divergence
between the posterior and prior, so that $\rt = -c(\xt)$ and $\rew_T =
D(\bel_T\|\bel_0)$ and in fact maximises the expected information
gain, while taking into account the cost of experimentation.

In this paper, an action corresponds to the choice of a specific drug
to test, and the outcome is the result of the test. We always use a
Gaussian process to model the distribution of outcomes $\yt$ given
drugs $\xt$. At any given time $t$, the process is denoted $\bt$.
However, maximising expected utility is intractable. In this paper, we
use a sparse lookahead to approximate the optimal solution, and we
show that this significantly outperforms other approximations, such as
Thompson sampling and upper confidence bound policies.

\section{Related work}
\label{sec:related-work}
Bandit problems are one of the most classical problems in resource allocation. For finite armed problems~\citep{Lai+Robbins} showed that the regret must grow logarithmically in the number of trials.
Subsequently, \citet{burnetas1996optimal} proved the existence of index-based optimal adaptive policies and constructed examples for e.g. distributions with finite support, while~\citet{mach:Auer+Cesa+Fischer:2002} constructed optimal policies for bounded distributions. However, in our case the allocation problem is budgeted~\citep{COLT:Madani:Budgeted:2004}, and so we are not interested in the cumulative regret. Specifically, our horizon is so short compared to the number of arms that many times not all arms are explored.

To handle the budget in our work, we define a fixed cost for each trial. However, one could also use an information-based stopping condition, as explored by~\citet{Burnetas+Katehakis:AsymptoticBayesBandit}. In that case the algorithm could decide for itself when a sufficient amount of information has been attained from the trials and could then terminate. While this is the right thing to do if the objective is to find the best arm, our cost constraints do not favour such an approach.

There has also been a lot of work on the connection between GPs and bandits. \citet{srinivas:gp-bandits:icml2010} combined the Upper Confidence Bound policy with Gaussian processes to obtain the GP-UCB policy. 
This policy uses the mean $\mu_x$ and variance $\sigma_x$ for each bandit context $x$ in the GP $\bel_t$ and sum them together.
It then selects and plays the bandit that maximizes this sum as per $x \in \argmax\limits_{x \in X} \mu_x + \sqrt{\beta_t} \sigma_x$. 
The variance is scaled by a factor $\sqrt{\beta_t}$ that varies with time and is used to handle the exploitation-exploration trade-off. 
This factor is set to be $\beta_t = 2\log (|D| t^2 \pi^2 / 6\delta)$ in this work since this works well if $D$ is finite, which it is for the data sets used in this work. 
After observing the result of playing the selected bandit the current belief is updated by letting the GP learn the new point as $\bel_{t+1} = \cb{t}{\xt,\yt}$.
~\cite{srinivas:gp-bandits:icml2010} use this policy to try to find the most congested part of a highway.

In some cases, a linear model might be sufficient to represent the
reward of different arms. Recently, linear bandits with Thompson
sampling (LB-TS) were used by ~\cite{agrawal2012:CoRR} for the
contextual bandit problem.  Their policy keeps track of the mean
$\hat{\mu}$ of the observed contexts and samples another mean
$\tilde{\mu} \sim \Normal(\hat{\mu},v^2 B^{-1})$.  $v$ is a scaling
factor used to make certain that the exploitation-exploration
trade-off is handled well and is set to
$v = R\sqrt{\frac{24}{\epsilon}d\ln\frac{1}{\delta}}$.  The $R$ is
data dependent and $\delta$ is an algorithm parameter.  $B$ is simply
a matrix of the contexts, $B = x_{t,i} x_{t,i}^T$.  The bandit that is
selected is $i \in \argmax_i x_{t,i}^T \tilde{\mu}$. We can apply
their algorithm in our problem; and since the horizon $T$ is known
beforehand we can set $\epsilon = \frac{1}{\ln T}$ as
in~\cite{agrawal2012:CoRR}.

The setting we consider, is similar to that of
\cite{king2004functional}. However it also has a number of other
challenges, mainly to do with the large dimensionality of $\Act$,
which is a finite subset of a high dimensional Euclidean space. 

To combat the curse of dimensionality, we use compressed
sensing~\citep{carpentier2012bandit} to make high-dimensional feature
space smaller.  This effectively means way multiplying the input
matrix data with a lower dimensional Gaussian matrix.  This maintains the distances
between the points with high probability and does not have a bad effect on performance.

Tree search methods for bandit problems has previously been explored
by \citet{wang:bayesian-sparse-sampling:icml:2005}, who proposed
Bayesian sparse sampling (BSS) as an action selection method. In their
paper, they compare BSS to more traditional methods such as
$\epsilon$-greedy, Boltzmann exploration and Thompson sampling.  Their
results show that BSS outperforms all other methods by a significant
margin using a GP model, at least for low-dimensional problems. 

\paragraph{Our contribution.} 
In this work, we explore two different tree search methods for the case of when there are numerous actions embedded in a high-dimensional space. 
We are particularly interested in the drug development application, wherein action selection is batch, because it is significantly cheaper and faster to test many drug compounds at the same time. 
For this reason, we develop a Thompson ranking algorithm and integrate it within tree search. We also show that batch actions have a significant side benefit. 
They very effectively deepen the search horizon while reducing the branching factor of the search tree, and consequently have a better performance than purely sequential methods.

\section{Sparse GP-tree lookahead policy}
\label{sec:sparse-gp-tree}
In this paper, we approximate the optimal adaptive experiment design
through sparse lookahead tree search. The tree is constructed in the
space of possible future beliefs, with the root node being the current
belief. All the beliefs are expressed as Gaussian process. At each
stage of the tree, we define the reward in a way that corresponds to
the problem definition, as specified in Section~\ref{sec:setting}. This then a Markov decision process (MDP), with a state space corresponding
to the set of possible information states (i.e. beliefs). Since there
set of such states is unbounded, we employ sampling-based
approximations to make planning tractable.

Let us start by defining the value function of the exact MDP. If
$\bt$ is our belief at some node at depth $t$ of the tree, 
the value function (i.e. the utility of the optimal policy) is:
\begin{equation}
\Vt(\bt) = \max_\pi \E_{\bt}(U_t \mid \pi).
\end{equation}

Via backwards induction, we can define the following recursion for
calculating the value function
\begin{align}
\Vt(\bt) &= \max_{\act \in \Act} \Qt(\bt, \act)\\
\Qt(\bt, \act) &= \int_\Obs \left\{ \rt(\bt, \act, \obs) + V_{t+1}(\cb{t}{\act, \obs}) \right\} \pdf_{\bt}(\obs \mid \act) \dd \obs
\\
\VT(\bt) &= \rt(\bt)
\end{align}
where $\act \in \Act$ is the experiment we wish to perform and $\cb{t}{\act,
	\obs}$ is the belief $\bt$ conditioned on $\act, \obs$, i.e.
\begin{equation}
\label{eq:conditioned belief}
\cbt{\act,\obs}(\cdot) = \bt(\cdot \mid \act, \obs)
\end{equation}

The problem in the GP setting is that $X$ is very large and that
$Y$ is infinite in size. For that reason, we shall replace the
first step of the recursion with
\begin{align}
\hVt(\bt) &= \max_{\act \in \Act(\bt)} \hQt(\bt, \act)\\
\Act(\bt) &\sim \textrm{Thompson}^{\nThompson}(\bt)
\end{align}
where $\Act(\bt)$ is a set of $\nThompson$ Thompson samples from the process
$\bt$. This lets us focus on a few promising actions;
the same idea was used by \citet{wang:bayesian-sparse-sampling:icml:2005} to perform Bayesian sparse sampling.

We use Thompson sampling in two ways in this work. 
Thompson sample rank is used in the batch version of the policy to quickly find a set of candidates for a single function. 
Independent Thompson sampling without replacement is used in all cases where testing is done sequentially.
\paragraph{Thompson sample rank.}
By $X(\bt) \sim \textrm{Thompson}^{\nThompson}(\bt)$, and we mean the following process.
We sample $\param \sim \bt$ to obtain the function $f_\param$. We create a permutation $\Act^\param$ of candidate drugs $\Act$ and rank them so that $f_\param(\act^\param_i) \geq f_\param(\act^\param_j)$ for any $\act_i^\param, \act^\param_j \in \Act$ such that $i < j$. Then we select the top $\nThompson$ drugs according to the sampled $\param$, $X(\bt) = \{x_1, \ldots, x_\nThompson\}$.

\paragraph{Independent Thompson sampling without replacement.}
If we perform independent Thompson sampling, we write
$X(\bt) \sim \textrm{Thompson}^{\nThompson}(\bt)$, and we mean the process whereby: For each $x_k(\bt) \in X(\bt)$, with $k = 1, \ldots, \nThompson$, 
we draw an independent sample $\param_k \sim \bel_t$ and set $x_k(\bt) = \argmax_x \E_{\param_k} [f \mid x] $.

The second step of the process involves the integration, but this
is much simpler. We can simply approximate the integral by Monte
Carlo sampling:
\begin{align}
\hQt(\bt, x) &= \sum_{y \in Y(\bt)} \rt(\bt, x, y) + \hV_{t+1}(\cb{t}{x, y}),\label{eq:hqt}\\ 
Y(\bt) &\sim P^{\nMC}(y \mid x, \bel_t).
\end{align}

\section{Data}
\label{sec:data}
\begin{figure}[tb]
	\begin{center}
		\includegraphics[width=6cm]{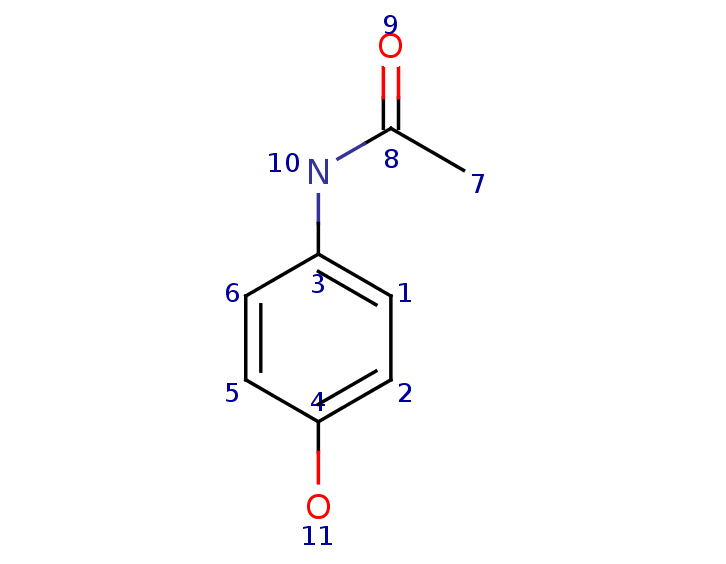}
	\end{center}
	\caption{Paracetamol is a drug molecule for treatment of pain and fever. The atom numbers are displayed in this figure.}
	\label{fig:paracetamol}
\end{figure}

The data used to test the performance of the algorithms developed in
this work consists of molecules in their graphical structure,
identified by their \textit{signature descriptors} as described
in~\citet{ci020345w}. An example of a molecule and corresponding signature descriptors are illustrated in Figure~\ref{fig:paracetamol} and Table~\ref{table:paracetamol}. These signature descriptors are then mapped into
high-dimensional space. 

\begin{table}[h!]
	\centering
	\begin{tabular}{rlll}
		\\
		\hline
		\\ [-2ex]
		\bf{Atom number} & \bf{Height $0$} & \bf{Height $1$} \\ [0.5ex]
		\hline
		\\ [-1.7ex]
		1 & \verb+[C]+ & \verb+[C]([C]=[C])+ \\
		2 & \verb+[C]+ & \verb+[C]([C]=[C])+ \\
		3 & \verb+[C]+ & \verb+[C]([C]=[C][N])+ \\
		4 & \verb+[C]+ & \verb+[C]([C]=[C][O])+ \\
		5 & \verb+[C]+ & \verb+[C]([C]=[C])+ \\
		6 & \verb+[C]+ & \verb+[C]([C]=[C])+ \\
		7 & \verb+[C]+ & \verb+[C]([C])+ \\
		8 & \verb+[C]+ & \verb+[C]([C][N]=[O])+ \\
		9 & \verb+[O]+ & \verb+[O](=[C])+ \\
		10 & \verb+[N]+ & \verb+[N]([C][C])+ \\
		11 & \verb+[O]+ & \verb+[O]([C])+ \\
		\hline
	\end{tabular} \vspace{.2cm}
	\caption{The signature descriptors of two different heights for paracetamol.}
	\label{table:paracetamol}
\end{table}
Each molecule also has its corresponding -log
IC$_{50}$ value. IC$_{50}$ denotes how much concentration is required
to inhibit a process by half. The datasets are patented molecules
retrieved from GOSTAR databases~\cite{gostar2012}. We have used two
different datasets; \textit{CDK5}, which describes different
molecules' inhibition of the cyklin-dependent kinase 5 (CDK5), and
\textit{MGLL}, which is other molecules' inhibition of the
monoacylglycerol lipase (MGLL).  The CDK5 data set comes with
IC$_{50}$ values in the range $[4.6,8.0]$.  The MGLL data set comes
with IC$_{50}$ values in the range $[4.8,8.4]$.  So each of the drugs
in the data sets come in the form of a $(x_n,y_{n}^*)$ pair where
$x_n$ is the nth bandit's context and $y_{n}^*$ is its corresponding
actual reward.

In addition to this, we generate a synthetic data set using the other two data sets by letting a 
GP learn the points in those data sets and then predict values for new, unknown points.

\begin{figure}
	\centering
%
%
\definecolor{mycolor1}{rgb}{0.00000,0.44700,0.74100}%
\definecolor{mycolor2}{rgb}{0.85000,0.32500,0.09800}%
\definecolor{mycolor3}{rgb}{0.92900,0.69400,0.12500}%
\definecolor{mycolor4}{rgb}{0.49400,0.18400,0.55600}%
\definecolor{mycolor5}{rgb}{0.46600,0.67400,0.18800}%
\definecolor{mycolor6}{rgb}{0.30100,0.74500,0.93300}%
\definecolor{mycolor7}{rgb}{0.63500,0.07800,0.18400}%
\begin{tikzpicture}[scale=0.5]

\begin{axis}[%
width=4.521in,
height=3.566in,
at={(0.758in,0.481in)},
scale only axis,
xmin=1,
xmax=100,
xlabel={Iteration (molecules tested)},
ymin=0.85,
ymax=1.25,
xlabel style={font=\LARGE},
ylabel={Avg cumulative regret},
ylabel style={font=\LARGE},
axis background/.style={fill=white},
title style={font=\bfseries,font=\LARGE},
title={CDK5},
legend style={legend cell align=left,align=left,draw=white!15!black,font=\Large}
]
\addplot [color=mycolor1,dashed,line width=2pt]
  table[row sep=crcr]{%
1	1.18465\\
2	1.199505\\
3	1.21612666666667\\
4	1.1647475\\
5	1.16124\\
6	1.14814666666667\\
7	1.13672\\
8	1.14173\\
9	1.15746888888889\\
10	1.143803\\
11	1.14187545454545\\
12	1.1308675\\
13	1.11575323076923\\
14	1.11267157142857\\
15	1.10190986666667\\
16	1.102768\\
17	1.10273282352941\\
18	1.09490983333333\\
19	1.08544015789474\\
20	1.08377865\\
21	1.08534585714286\\
22	1.08230377272727\\
23	1.07684795652174\\
24	1.07204083333333\\
25	1.0647026\\
26	1.05640053846154\\
27	1.05269425925926\\
28	1.04560142857143\\
29	1.04260617241379\\
30	1.0373179\\
31	1.03511461290323\\
32	1.03534946875\\
33	1.03309972727273\\
34	1.02995411764706\\
35	1.02874011428571\\
36	1.02749913888889\\
37	1.02273640540541\\
38	1.01751231578947\\
39	1.01591007692308\\
40	1.014546\\
41	1.01500731707317\\
42	1.01148095238095\\
43	1.00903411627907\\
44	1.00751665909091\\
45	1.00535344444444\\
46	1.00009347826087\\
47	0.998945510638298\\
48	0.997586041666667\\
49	0.993891795918368\\
50	0.99144974\\
51	0.987951568627451\\
52	0.985494346153847\\
53	0.983189132075472\\
54	0.981484333333334\\
55	0.980234090909091\\
56	0.977363142857143\\
57	0.977500789473684\\
58	0.976329293103448\\
59	0.975577203389831\\
60	0.973469316666667\\
61	0.971558573770492\\
62	0.971454096774194\\
63	0.968998476190476\\
64	0.96676471875\\
65	0.962874630769231\\
66	0.962218393939394\\
67	0.959439417910448\\
68	0.956892514705882\\
69	0.954154724637681\\
70	0.953282914285714\\
71	0.951550802816901\\
72	0.949112972222222\\
73	0.945617\\
74	0.944138054054054\\
75	0.943150346666667\\
76	0.941679381578947\\
77	0.939223168831169\\
78	0.936487179487179\\
79	0.934205\\
80	0.9312359625\\
81	0.92862924691358\\
82	0.926983792682926\\
83	0.924389240963855\\
84	0.923773988095238\\
85	0.921971352941176\\
86	0.920845197674418\\
87	0.919273735632183\\
88	0.91707275\\
89	0.914879516853932\\
90	0.912938766666666\\
91	0.913041362637362\\
92	0.910091065217391\\
93	0.909681741935483\\
94	0.90905829787234\\
95	0.906952389473684\\
96	0.903347249999999\\
97	0.902483690721649\\
98	0.901193091836734\\
99	0.898848282828282\\
100	0.897068479999999\\
};
\addlegendentry{GP-Tree Policy (CS)};

\addplot [color=mycolor2,dashed,line width=2pt]
  table[row sep=crcr]{%
1	1.08327\\
2	1.140035\\
3	1.19977666666667\\
4	1.1507525\\
5	1.150074\\
6	1.143875\\
7	1.13629285714286\\
8	1.14498\\
9	1.15399777777778\\
10	1.1362894\\
11	1.12430763636364\\
12	1.12024033333333\\
13	1.10413984615385\\
14	1.09898342857143\\
15	1.09435653333333\\
16	1.096058625\\
17	1.09076694117647\\
18	1.08455394444444\\
19	1.07163084210526\\
20	1.0705228\\
21	1.07519980952381\\
22	1.07301390909091\\
23	1.06864217391304\\
24	1.06363625\\
25	1.0562976\\
26	1.04851411538462\\
27	1.04615392592593\\
28	1.04168289285714\\
29	1.03928265517241\\
30	1.03551036666667\\
31	1.03116964516129\\
32	1.02955275\\
33	1.0297663030303\\
34	1.02642385294118\\
35	1.0277386\\
36	1.02571841666667\\
37	1.02256727027027\\
38	1.02249997368421\\
39	1.01954856410256\\
40	1.01738615\\
41	1.01801697560976\\
42	1.01749788095238\\
43	1.01529465116279\\
44	1.0171175\\
45	1.0149174\\
46	1.01271130434783\\
47	1.01024817021277\\
48	1.00456352083333\\
49	1.00277773469388\\
50	1.00062662\\
51	0.995984215686274\\
52	0.995897346153846\\
53	0.994121698113207\\
54	0.989855833333333\\
55	0.989334836363636\\
56	0.985224857142857\\
57	0.982995052631579\\
58	0.983714103448276\\
59	0.982067101694915\\
60	0.980245616666666\\
61	0.979386245901639\\
62	0.97943\\
63	0.976588333333333\\
64	0.97545859375\\
65	0.971405753846154\\
66	0.970322787878788\\
67	0.968663373134328\\
68	0.966235647058823\\
69	0.962873231884058\\
70	0.961349242857143\\
71	0.958884126760563\\
72	0.956655125\\
73	0.953625602739726\\
74	0.951836567567567\\
75	0.951217013333333\\
76	0.950246157894737\\
77	0.947105363636364\\
78	0.94460258974359\\
79	0.941730088607595\\
80	0.9390336875\\
81	0.93702762962963\\
82	0.935157158536585\\
83	0.932759228915663\\
84	0.931941011904762\\
85	0.930084164705882\\
86	0.927726313953488\\
87	0.926365954022989\\
88	0.924296193181818\\
89	0.921630146067416\\
90	0.919682233333333\\
91	0.919362802197802\\
92	0.917002336956522\\
93	0.916774731182796\\
94	0.915754936170213\\
95	0.913528042105263\\
96	0.911882416666667\\
97	0.910848072164949\\
98	0.909739357142857\\
99	0.907833454545455\\
100	0.90613957\\
};
\addlegendentry{GP-Tree Policy (NO CS)};

\end{axis}
\end{tikzpicture}%
	\caption{A comparison of the main algorithm ran on data pre-processed with Compressed sensing and without}
	\label{fig:CS}
\end{figure}
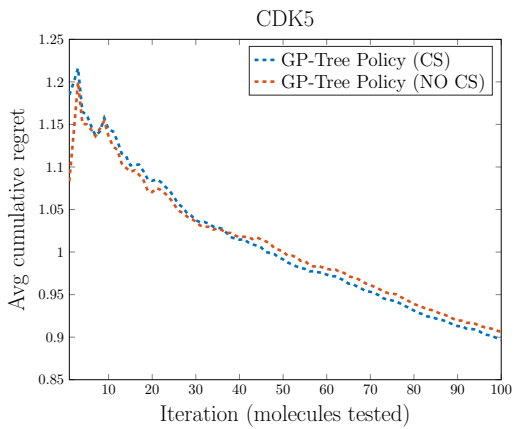

The data is compressed using Compressed sensing as mentioned in Section~\ref{sec:related-work}, 
the results in Figure~\ref{fig:CS} show that any accuracy loss compared to not using Compressed sensing is minor. 

The metrics used to compare the performance of the algorithms are 
the average regret (\ref{eq:aregret}) and the simple regret (\ref{eq:sregret}). 
For these data sets we use definition (a) definition for the utility function and let 
the actual reward be $r_{i,t} = y_{i}^*$. 
Let $r^{*}_t = \max_i y_{i}^*$.
We then define the (instantaneous) regret to be the difference from 
the reward $r_{i,t}$ and the optimal reward $r^{*}_t$.
Since the rewards are fixed for each drug this can be simplified to 
\begin{align}
\iregret = r^{*}_t - r_{i,t}. \label{eq:iregret}
\end{align}

The average regret is used to compare how well the 
algorithms manage to identify and select the best drugs in the data set.
\begin{align} 
\aregret = \sum_{t=1}^{T} \frac{r^{*} - r_{i,t}}{T}. \label{eq:aregret}
\end{align}

The simple regret is used to compare how well the algorithms manage 
to identify and select the best drug in the data set.
\begin{align}
\sregret = r^{*} - r_{i,t}. \label{eq:sregret}
\end{align}

Internally, the reward measures when optimizing for $\mathbb{E}\sregret$ are slighty different than from when we optimize for $\mathbb{E}\aregret$. 
All intermediary nodes have their reward set to zero as we only care about the reward at the end. 
So Equation~\ref{eq:hqt} becomes the following instead,
\begin{align}
\hQt(\bt, x) &= \sum_{y \in Y(\bt)}\hV_{t+1}(\cb{t}{x, y}).
\end{align}

The two regret measures describe two significantly different optimization strategies of the problem. 
What is most desired depends on the context but generally it is desired to minimize $\mathbb{E}\sregret$ in the case 
where the single most promising drug is to be discovered and $\mathbb{E}\aregret$ where the goal is to identify multiple 
promising drugs.
The work by~\citet{bubeck2011pure} show that a single policy can not guarantee optimal $\mathbb{E}\sregret$ and $\mathbb{E}\aregret$ 
bounds at the same time so each policy is run twice, each with one of the optimization goals in mind.

\section{Algorithms}
\label{sec:algorithms}

The GP-Thompson Policy, as shown in Algorithm (\ref{alg:gp-ts}) is used to test whether there is a 
increase in performance or not by using the lookahead. 

\begin{algorithm}
	\caption{GP-Thompson Policy}
	\label{alg:gp-ts}
	\begin{algorithmic}
		\STATE \textbf{input} prior $\bel_0$, actions $X$
		\FORALL{$t=1,2,...,T$}
		\STATE Play arm $\xt \in \argmax\limits_{x \in X(\bt)} f(x) \sim \bt$
		\STATE Observe outcome $\yt$ and reward $\rt$
		\STATE Update belief $\bel_{t+1} = \cb{t}{\xt,\yt}$.
		\ENDFOR
	\end{algorithmic}
\end{algorithm}
The algorithm approximates the rewards that would be acquired by taking each of 
the actions in turn by using Thompson sampling. It then selects and 
carries out the action that is deemed the best from the sampling. 
The actual reward is then observed and learned by the GP.
The $\bel_0$ parameter is the prior belief before any actions have been taken and rewards observed. 
$\bt$ reflects policy's state of belief at time $t$ and $X$ the set of possible actions to carry out.

Algorithm (\ref{alg:gp-tree}) describes the main policy used in this work. 
The policy is further expanded in Algorithm (\ref{alg:batch-gp-tree}) to handle batch 
testing. However, initially we just consider sequential testing with one action being taken 
at each time step.

\begin{algorithm}
	\caption{GP-Tree Policy}
	\label{alg:gp-tree}
	\begin{algorithmic}
		\STATE \textbf{input} horizon $h$, branches $K$, samples $n$, prior $\bel_0$, actions $X$, depth $\depth$
		\FORALL{$t=1,2,...,T$}
		\STATE Play arm $x \in \argmax\limits_{x \in X(\bt)}$ \textsf{descendQ}($\bt$,$x$,$\depth=0$,$h$,$K$)
		\STATE Observe outcome $\yt$ and reward $\rt$
		\STATE Update belief $\bel_{t+1} = \cb{t}{\xt,\yt}$.
		\ENDFOR
	\end{algorithmic}
\end{algorithm}
\begin{algorithm}
	\caption{descendQ}
	\label{alg:descend-Q}
	\begin{algorithmic}
		\STATE \textbf{input} horizon $h$, branches $K$, samples $n$, GP $\bt$, actions $X$, action $x$, depth $\depth$
		\FORALL{$k=1,2,...,K$}
		\STATE $y_{t,k} \leftarrow f(x) \sim \bt$
		\STATE Update belief $\bel_{t+1,k} = \cb{t}{\xt,y_{t,k}}$.
		\STATE $\hat{Q}_{t,k}(\bel_{t+1,k}, x) = $ \textsf{descendV}$(\bel_{t+1,k},x,\depth + 1,h,K)$
		\ENDFOR
		\STATE \textbf{return} $\sum_{k=1}^{K} \frac{\hat{Q}_{t,k}(\bel_{t+1,k}, x)}{k}$
	\end{algorithmic}
\end{algorithm}
\begin{algorithm}
	\caption{descendV}
	\label{alg:descend-V}
	\begin{algorithmic}
		\STATE \textbf{input} horizon $h$, branches $K$, samples $n$, GP $\bt$, actions $X$, action $x$, depth $\depth$
		\IF{$\depth = h$}
		\STATE \textbf{return} $\max\limits_{x \in X} f(x) \sim \bt$
		\ELSE
		\STATE \textbf{return} $\max\limits_{x \in X} $ \textsf{descendQ}($\bt$,$x$,$\depth$,$h$,$K$)
		\ENDIF
	\end{algorithmic}
\end{algorithm}

The GP-Tree Policy is built on the same idea as the GP-Thompson policy, however, 
we now also consider the non-myopic effects of taking the actions.  
We select and carry out an action and then we let our current belief state $\bt$ learn the 
predicted reward of taking that action. We can then move to a new belief state $\bel_{t+1}$ and 
from there on select and take the actions using $\bel_{t+1}$ instead of $\bt$. 
This process is then continued, sampling and carrying out $n$ actions at each time step as well as 
branching our belief state into $K$ several belief states $\bel_{t,1},\bel_{t,2},...,\bel_{t,K}$ each with their 
predicted reward $y_{t,1},y_{t,2},...,y_{t,K}$, until the horizon $h$ is reached which is when the predicted accumulated rewards in the tree 
will propagate back to our actual time step. 
The policy is then able to take the action that is not only the most promising at this moment but also the one 
that will give the best results in the future, as predicted by $\bt$. 
So in fact we build a balanced search tree of width $nK$ and height $h$.

\begin{algorithm}
	\caption{Batch GP-Tree Policy}
	\label{alg:batch-gp-tree}
	\begin{algorithmic}
		\STATE \textbf{input} horizon $h$, branches $K$, samples $n$, prior $\bel_0$, actions $X$, depth $\depth$, batch size $b$
		\FORALL{$t=1,2,...,\frac{T}{b}$}
		\STATE Play arms $\boldsymbol{x} \in \argmax\limits_{\boldsymbol{x} \in X(\bt)}$ \textsf{descendQ}($\bt$,$\boldsymbol{x}$,$\depth=0$,$h$,$K$)
		\STATE Observe outcomes $\boldsymbol{y}_t=\yt, y_{t+1},...y_{t+b}$ and rewards $\boldsymbol{r}_t=\rt,r_{t+1},...r_{t+b}$
		\STATE Update belief $\bel_{t+b} = \cb{t}{\boldsymbol{r}_t,\boldsymbol{y}_t}$.
		\ENDFOR
	\end{algorithmic}
\end{algorithm}

One of the main drawbacks with this policy is that it is fairly computationally expensive 
since the number of nodes to explore grows rapidly with further horizons. 
To combat this we extend the GP-Tree Policy to Batch GP-Tree Policy as can be seen in Algorithm (\ref{alg:batch-gp-tree}). 
The idea is the same however instead of taking a single action $x$ at time step $t$ we take $b$ actions at the same time. 
This means that we expand the action space from $x \in X$ to $\boldsymbol{x} \in X^b$. 
The reasoning for this is that we do not have to consider the effect on the belief of taking every action sequentially and instead how a series of actions 
affect the belief state. This is also how drug testing in experiment design is done in reality.
By doing this change we can have higher values on our $h$, $K$, $n$ parameters compared to the GP-Tree Policy and still have the policy finish in reasonable time.

\section{Results}
\label{sec:results}
\begin{figure}
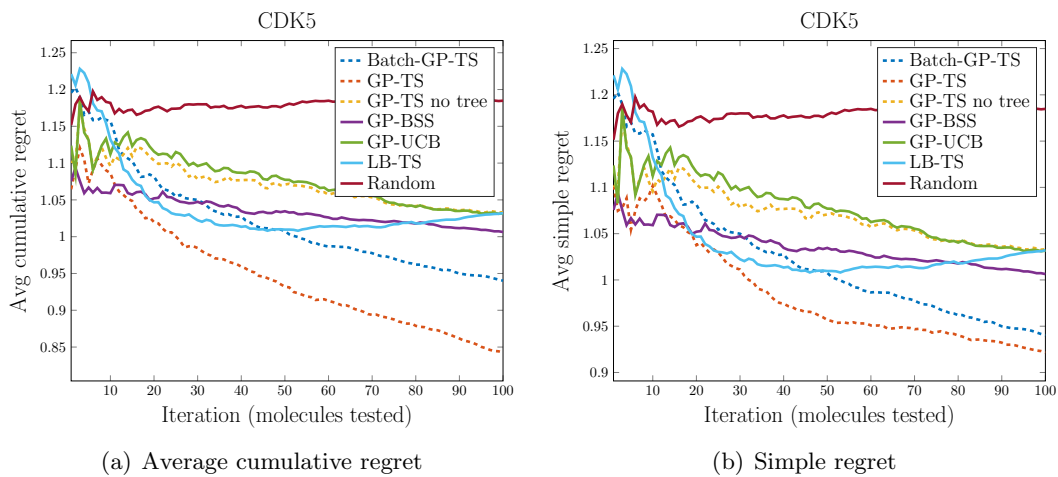

	\centering
	\subfigure[Average cumulative regret]{
		\input{CDK_c.tex}
		\label{fig:CDK_c}
	}
	\subfigure[Simple regret]{
		\input{CDK_s.tex}
		\label{fig:CDK_s}
	}
	\caption{Results on the CDK5 data set)}
	\label{fig:CDK5}
\end{figure}

\begin{figure}
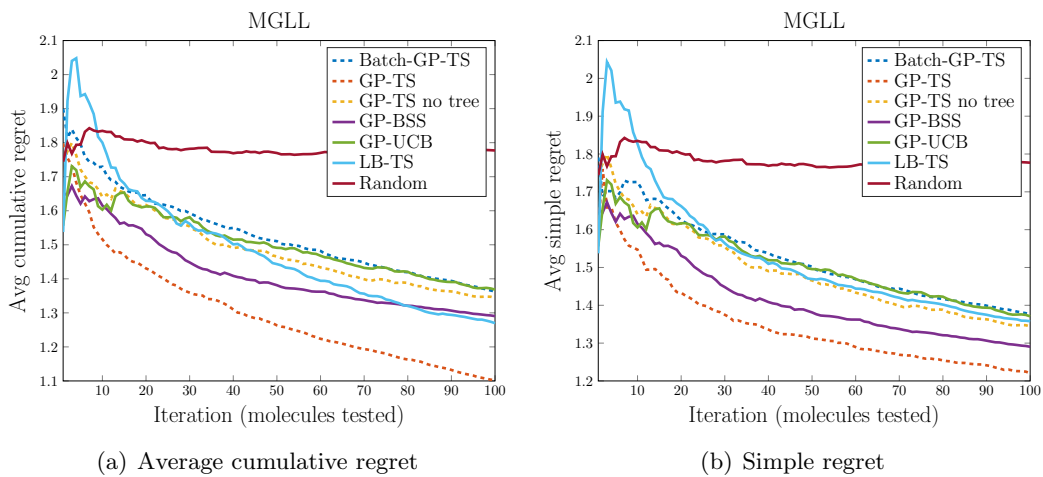

	\centering
	\subfigure[Average cumulative regret]{
		\input{MGLL_c.tex}
		\label{fig:MGLL_c}
	}
	\subfigure[Simple regret]{
		\input{MGLL_s.tex}
		\label{fig:MGLL_s}
	}
	\caption{Results on the MGLL data set}
	\label{fig:MGLL}
\end{figure}

\begin{figure}
	\centering
	\subfigure[Average cumulative regret]{
		\input{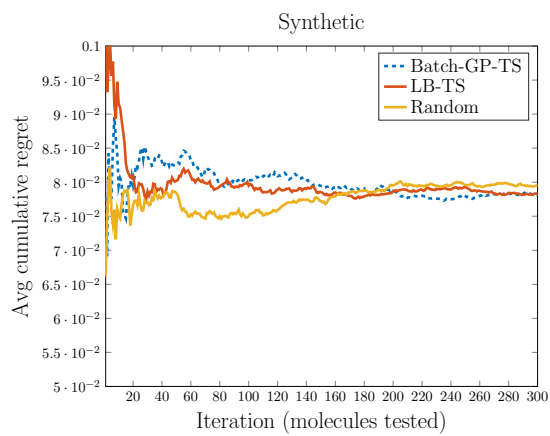}
		\label{fig:synt_c}
	}
	\caption{Results on the synthetic data set}
	\label{fig:synt}
\end{figure}

The methods are run on three different data sets, one small (CDK5, see Figure~\ref{fig:CDK5}), one large (MGLL, see Figure~\ref{fig:MGLL}) and one huge (synthetic, see Figure~\ref{fig:synt}). 
In all the runs on the three data sets, information about one drug is initially shown and then for the remaining $T-1$ iterations the policies have to decide for themselves which drugs should be tested. 
Results given by the main policies studied in this work have dashed lines and those that they are compared to are in solid lines. 

The algorithm parameters were obtained through empirical testing on firstly, a small function optimization problem and later on a small drug data set. 
For policies that are time expensive, most notably all tree search policies, had their parameters tuned in such a way so the evaluation of all of them takes 
about the same amount of time. 
The methods may have one set of parameters for each optimization goal.

Some settings are consistent throughout all the runs and they will be gone over here.
The problem horizon $T$ is the limit of the number of molecules that can be tested in a single experiment and is set as $T=100$ for all runs. 
The Gaussian processes $\bel$ that are used are using a RBF kernel to measure similarity between contexts and $\bel$ has internal signal noise and noise variance of $0.1$. 
GP-UCB uses $\beta_t = 2\log (|D| t^2 \pi^2 / 6\delta) $, as in ~\cite{srinivas:gp-bandits:icml2010}.

Next are the data specific settings. As a result of the sheer size of the two larger data sets, these parameters have to be tuned carefully in order for the methods to evaluate in 
reasonable time and also to give the methods a chance to learn the data.
Batch GP-Tree Policy is run with the following configuration on the CDK5 data set, $b = 5$, $n = 100$, $K = 4$, $h = 1$. 
GP-Tree Policy is run with the same settings as in the batch case apart from the number of Thompson samples, which is $n=20$ in this case.
The GP-BSS Policy is run with the parameters $\textsf{budget}=800$, $\rho=0.5$.
The GP-UCB Policy and LB-TS Policy both have a separate $\delta$ depending on the optimization goal, here $\delta_{\sregret} = 0.99$, $\delta_{\aregret} = 0.01$ when optimizing for $\mathbb{E}\sregret$ and $\mathbb{E}\aregret$, respectively.
The results when optimizing for $\mathbb{E}\aregret$ are presented in Figure~\ref{fig:CDK_c}. 

As the MGLL data set is much larger than the CDK5 data set, some parameters are now changed to speed up the evaluation. 
These are $b=10$ for Batch GP-Tree Policy, $n=10$ for GP-Tree Policy and $\textsf{budget}=400$ for GP-BSS Policy. 
The results when optimizing for $\mathbb{E}\aregret$ are presented in Figure~\ref{fig:MGLL_c}. 

Finally, since the amount of data available in the data sets is relatively small, we also worked with a larger, synthetic data set. 
This was created by training a GP model on the complete data, and then generating samples from the GP for 3000 points.
All the GP based policies are too slow evaluate on this data set apart from the Batch GP-Tree Policy with sufficient batch size. 
The settings on the synthetic data set have the following changes, $b=200$, $n=40$ and $K=2$.
Results are shown in Figure~\ref{fig:synt_c} when optimizing for $\mathbb{E}\aregret$.

\section{Conclusions}
\label{sec:future-work}

We have shown that the two tree search methods work well in a setting with a large amount of actions in high-dimensional space. 
In particular, we have shown that the policies are competitive or work better than similar action selection policies, at least in the case of high-dimensional experiment design. 
We demonstrated the need and effectiveness of batch action selection in this setting when the number of actions is scaled up immensely.
With the Batch GP-Tree Policy we can take advantage of the prediction accuracy of the tree search at the same time as we can keep the computational complexity low enough for the 
problem to be solved in reasonable time. 

Future work and further improvements of the two tree search policies will be discussed hereafter.

Empirically proving tight $\mathbb{E}\sregret$ bounds. This would require extensive testing since the simple regret of these policies inherently have high variance. 
Through further testing it would be possible to construct confidence bounds on the simple regret.

Extending the support for batch testing by adaptively changing the batch size depending in some way on how confident the policy is in its predictions. 
That way it would be possible for the policy a situation when it inevitably has to take some bad actions just to fill out the rest of the batch.

Adding support for multi-objective optimization by finding the Pareto optimal arms. 
Similar work has been done in~\citet{durandimproving} for UCB1 and in~\citet{yahyaathompson} for Bernoulli bandits.

Adding support for selection over multiple experiments. 
This could be interesting when there are multiple different labs available to test the drugs. 
Perhaps some of them are more accurate than others and also have varying costs and durations. 
The policy should also preferably learn from all the experiments and tests which may lead to a 
complex situation with many GPs depending on each other. Perhaps the results in~\citet{boyle2004dependent} could be used for this.

An interesting idea would be to have an adaptive bandit that could 
change the branching factor and the horizon depending on the 
number of trials that has been carried out so far. Perhaps having a 
deeper search tree in the beginning and a wider search tree in the end 
could lead to greater results.

Another idea is to replace the current Compressed sensing algorithm 
with a differentially private one as in~\citet{Li:2011:CMU}.

\acks{This work is supported by a Chalmers Information and Communication Technology (ICT) Area of Advance (AoA) SEED (2015--2016) grant.}

\vskip 0.2in
\bibliography{misc,christos}

\end{document}